\documentclass[letterpaper, 10 pt, conference]{ieeeconf}  

\usepackage{microtype}
\usepackage{graphics}
\usepackage{subcaption}
\usepackage{booktabs} 

\usepackage{hyperref}
\usepackage{cite}

\usepackage{makecell}

\usepackage{times}  
\usepackage{helvet} 
\usepackage{courier}  

\usepackage{epstopdf}
\usepackage{graphicx}
\usepackage{rotating,multirow}
\usepackage{amsmath,amssymb} 
\usepackage{color}
\usepackage{wrapfig,lipsum}
\usepackage{tabularx}

\usepackage{algorithm}
\usepackage{algorithmic}
\usepackage[algo2e]{algorithm2e}
\usepackage{mathtools}

\newcommand{\etal}{\textit{et al. }}
\newcommand{\eg}{\textit{e.g., }}
\newcommand{\ie}{\textit{i.e., }}
\newcommand{\aka}{\textit{a.k.a. }}

\IEEEoverridecommandlockouts                              

\title{\LARGE \bf Unsupervised Vehicle Re-Identification via Self-supervised Metric Learning using Feature Dictionary}

\author{Jongmin Yu$^{1,\dagger}$ and Hyeontaek Oh$^{1}$
\thanks{$\dagger$ denotes the corresponding author.}
\thanks{$^{1}$Institute for IT Convergence, Korea Advanced Institute of Science and Technology (KAIST), Daejeon, Republic of Korea (34141)
        {\tt\small \{andrew.yu, hyeontaek\}@kaist.ac.kr}}
}

\begin{document}

\maketitle
\thispagestyle{empty}
\pagestyle{empty}

\begin{abstract}
The key challenge of unsupervised vehicle re-identification (Re-ID) is learning discriminative features from unlabelled vehicle images. Numerous methods using domain adaptation have achieved outstanding performance, but those methods still need a labelled dataset as a source domain. This paper addresses an unsupervised vehicle Re-ID method, which no need any types of a labelled dataset, through a Self-supervised Metric Learning (SSML) based on a feature dictionary. Our method initially extracts features from vehicle images and stores them in a dictionary. Thereafter, based on the dictionary, the proposed method conducts dictionary-based positive label mining (DPLM) to search for positive labels. Pair-wise similarity, relative-rank consistency, and adjacent feature distribution similarity are jointly considered to find images that may belong to the same vehicle of a given probe image. The results of DPLM are applied to dictionary-based triplet loss (DTL) to improve the discriminativeness of learnt features and to refine the quality of the results of DPLM progressively. The iterative process with DPLM and DTL boosts the performance of unsupervised vehicle Re-ID. Experimental results demonstrate the effectiveness of the proposed method by producing promising vehicle Re-ID performance without a pre-labelled dataset. The source code for this paper is publicly available on \url{https://github.com/andreYoo/VeRI_SSML_FD.git}.
\end{abstract}

\section{Introduction}
\label{sec:intro}
Vehicle re-identification (Re-ID) aims to find the same vehicle image for a given probe vehicle image from a vehicle image database captured from various cameras. Vehicle Re-ID has been achieved a great number of successes alongside the development of deep learning \cite{he2016deep,huang2017densely}. Notably, the extraordinary feature extraction capacity of supervised deep learning for a large-scale and well-labelled dataset significantly improves vehicle Re-ID performances. However, it is hard and expensive to create a large-scale and well-labelled dataset. Moreover, mistakenly labelled images on the dataset can degrade the vehicle Re-ID performance of those supervised methods. As a result, attention to unsupervised methods has been only increasing in recent, which do not require any pre-labelled dataset.

Learning discriminative representation without a pre-labelled dataset is one of the important challenges for unsupervised vehicle Re-ID. Domain adaptation (DA), which transfers informative features of other labelled datasets (\aka source domains) to an unlabelled dataset (\aka target domain), is the predominant approach to solve this challenge \cite{peng2020unsupervised,huang2020dual,he2020multi}. It has been comprehensively applied for various object identification problems (\eg face identification \cite{sohn2017unsupervised,hong2017sspp} and person Re-ID \cite{deng2018image,liu2019adaptive,yang2020part}). Those approaches have remarkably improved the performance of vehicle Re-ID \cite{he2019part,zhong2019invariance,song2020unsupervised}.

\begin{figure}[t]
	\centering
	\includegraphics[width=\columnwidth]{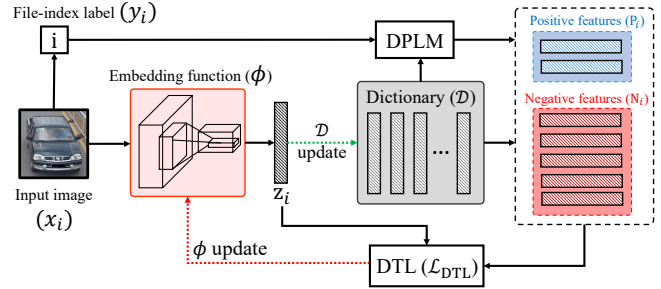}
	\caption{An overview of the proposed self-supervised metric learning (SSML) for unsupervised vehicle Re-ID using the proposed dictionary-based positive label mining (DPLM) and the proposed dictionary-based triplet loss (DTL). DPLM and DTL complementarily work to boost the vehicle Re-ID performance without a labelled dataset. DPLM provides positive and negative labels, and DTL improves the performance of the embedding function using the labels, iteratively.}
	\label{fig:1}
	\vspace{-2ex}
\end{figure}

However, DA based methods can only avoid the requirement of labelling jobs about the target domain, and they still need a labelled dataset from a source domain. Additionally, a domain gap, which can be measured by probabilistic or geometric difference on learnt features between the source domain and the target domain, is also a substantial issue in applying DA for a Re-ID task \cite{wei2018person}. It is not only unpredictable but also hard to measure before evaluating the model performances \cite{wang2020unsupervised}. 

On studies about person Re-ID, to develop fully unsupervised approaches for Re-ID, in recent few years, generating pseudo labelling using clustering methods \cite{fan2018unsupervised,Fu_2019_ICCV} or computing pair-wise similarity (PS) between features \cite{wang2020unsupervised,lin2020unsupervised} have been presented. However, the label quality is variant to hyper-parameter settings (\eg the number of clusters or initial positions of centroids). A fully unsupervised approach for vehicle Re-ID is still an unexplored area.

For a fully unsupervised vehicle Re-ID, this paper presents self-supervised metric learning (SSML) to improve the performance without any kinds of labelled data. As shown in Fig. \ref{fig:1}, the proposed SSML initially extracts feature from a given image and assigns a single-class label (\ie the file indexes). This job is conducted for all training images. Then, SSML generates a dictionary to store all features using the labels as keys. Thereafter, SSML conducts the proposed dictionary-based positive mining (DPLM) to find images marked positive labels that may belong to the same vehicle with an input image. The PS between features, the relative-rank consistency, and the adjacent feature distribution similarity are jointly considered for finding the features. The excavated positive labels are applied to the proposed dictionary-based triplet loss (DTL) for optimising our method.

SSML using DPLM and DTL allows applying metric learning for a large-scale unlabelled dataset under an unsupervised manner. This approach can reduce the complexity of our method because it is not required to create a fully-connected network for classifying vehicle identities. The dictionary is constructed as a non-parametric function, so it will not affect the entire model complexity. Additionally, DTL includes a hard-sampling mining task to address the quantitative unbalance between positive and negative samples. As a result, SSML with DPLM and DTL provides a stable optimisation approach, which invariant to the scalability of a dataset, to improve the performance of unsupervised vehicle Re-ID.

Our model is evaluated on VeRi-776 dataset \cite{liu2016deep} and VeRi-Wild dataset \cite{wei2018person}. Compared with recent state-of-the-art unsupervised vehicle Re-ID methods, including DA based methods, our method shows competitive performance or outperforms them. Our method produces rank-1 accuracy of 74.5 on VeRi-776 which is better than SSL \cite{lin2020unsupervised} and MMLP \cite{wang2020unsupervised}. In comparison with DA based unsupervised vehicle Re-ID methods, our method shows comparable performance without a labelled dataset as a source domain. 
Consequently, the proposed method achieves promising performance without any types of labelled data.

\section{Related Works}
\label{sec:rw}
\subsection{Supervised Vehicle Re-identification}
The majority of the vehicle Re-ID methods is derived based on a supervised manner. In recent decades, with the development of outstanding feature learning methods based on deep learning and the dissemination of various large-scale vehicle Re-ID datasets \cite{liu2016deep,liu2016deep_db,wei2018person}, supervised vehicle Re-ID has been improving remarkably. In particular, the outstanding feature learning based on Convolutional Neural Networks (CNNs) boosts vehicle Re-ID performance rapidly \cite{wang2017orientation,spatio_temporal_path,liu2016deep}. Wang \etal \cite{wang2017orientation} proposed a feature learning approach based on an orientation-based region proposal. Shen \etal \cite{spatio_temporal_path} leveraged learnt features from CNNs to aggregate spatio-temporal information. Recently, various methods applying meta-information (such as vehicle parts, key-points segmentation \cite{sensetime17}, or view-point awareness \cite{zheng2019attributes,viewpoint_aware}) have been proposed to improve the discriminative power of vehicle Re-ID methods.

Metric learning is also one of the frequently used approaches for the supervised vehicle Re-ID \cite{gste,antonio2018unsupervised}. Bai \etal \cite{gste} proposed a deep metric learning method called group-sensitive-triplet embedding (GS-TRE) to recognise and retrieve vehicles. Antonio \etal \cite{antonio2018unsupervised} leveraged pairwise and triplet constraints for training a network that is capable of assigning a high degree of similarity to samples with the same identity. Those metric learning-based methods just need annotations about whether two images contain the same vehicle or not. However, even though the label is softened, it remains a problem that prepared labels are needed to train their models, and their performances highly depend on the label quality. Consequently, it is necessary to develop a method that is free to pre-labelled datasets.

\subsection{Unsupervised Vehicle Re-identification}
Compared with the vehicle Re-ID with supervision, the studies on unsupervised vehicle Re-ID is in the beginning because it is intractable to learn discriminative features without labels. To tackle this obstacle, it has been proposed that methods \cite{huang2020dual,he2020multi,peng2020unsupervised} to leverage annotated information of other pre-labelled datasets (\aka domain adaptation (DA)). Huang \etal \cite{huang2020dual} presented a dual-domain multi-task model that divides the vehicle images into two domains based on a frequency. He \etal \cite{he2020multi} proposed a multi-domain learning method to jointly consider the real-world data and synthetic data for model training. Peng \etal \cite{peng2020unsupervised} presented a data adaptation module generating the images with similar data distribution to the unlabelled target domain. However, those methods need a labelled source dataset; that is, those are still can not be applied when a model can only use a non-labelled dataset. Therefore, it is essential to develop an unsupervised method to overcome the necessity of source domains. 

Compared with the studies using DA, regrettably, fully unsupervised vehicle Re-ID is not explored yet. However, in the literature of person Re-ID, several approaches have been proposed to address a fully unsupervised way \cite{fan2018unsupervised,Fu_2019_ICCV,wang2020unsupervised,lin2020unsupervised}.  Fan \etal \cite{fan2018unsupervised} and Fu \etal \cite{Fu_2019_ICCV} presented clustering-based unsupervised person Re-ID methods. Wang and Zhang \cite{wang2020unsupervised} and Lin \etal \cite{lin2020unsupervised} proposed pseudo-label generation methods to find person images containing the same identity. Those methods are variant to the hyper-parameter settings such as the number of clusters or the initial position of centroids.
In this paper, we propose an unsupervised vehicle Re-ID method that reduces the dependency of these hyper-parameters. To the best of our knowledge, our method is the first method to address fully unsupervised vehicle Re-ID issue, which does not need prior preparation of labelled dataset.

\section{Self-supervised Metric Learning}
\label{sec:pm}
\subsection{Methodology Overview}
The goal of the proposed method is compiling a discriminative vehicle Re-ID model using unlabelled vehicle images $\mathcal{X}=\{x_{i}\}^{i=1:n}$, where $n$ is the number of images. To achieve this, the proposed method is designed based on a self-supervised approach. As shown in Fig. \ref{fig:1}, at the beginning of the training, a single-number label $y_{i}$, defined by file index orders, is assigned to each image $x_{i}$ as $y_{i} = i$. Even though there are multiple images showing the same vehicle, each image has an independent label at this moment. Above the file index-wise labelling method is inspired from person Re-ID methods \cite{wang2020unsupervised,lin2020unsupervised}. After that, the proposed method extracts the features ${\mathcal{Z}}$ from all training images in $\mathcal{X}$. An embedding function $\phi$ extracts a latent feature $z_{i}$ from an image $x_{i}$ as $\phi$: $x_{i}\longrightarrow{}z_{i}$, $z_{i}\in\mathbb{R}^{d}$, where $d$ is the output dimension of $\phi$. The extracted feature $z_{i}$ is regularised by $l_{2}$-normalisation, so the scale of all features is fixed by 1. The extracted features $\mathcal{Z}=\{z_{i}\}^{i=1:n}$ are applied to create and update a dictionary $\mathcal{D}=[\bar{z}_{i}]^{i=1:n},{  }\mathcal{D}\in\mathbb{R}^{n\times{}d}$, where $\bar{z}_{i}$ is a corresponding feature for $x_{i}$ and $n$ is the number of dictionary elements in which the same as the number of training images. 

During the training step, with a given unlabelled image $x_{i}$, the proposed method searches a positive label set and a negative label set using the dictionary $\mathcal{D}$. The positive label set consists of labels of images that may belong to the same vehicle identity to $x_{i}$, and all other labels would be considered as a negative label. The proposed method conducts metric learning with those sets. In this workflow, it is essential to mine flawless reliable positive and negative label sets. 

The most straightforward approach, to find a positive label set for a given image $x_{i}$, is computing PS across entire training images. In our method, it can be obtained by computing PS between $z_{i}$ and $\mathcal{D}$ as follows:
\begin{equation}
\begin{aligned}
s_{i} = \phi(x_{i})\mathcal{D}^{\mathrm{T}}&=[z_{i}\bar{z}^{\mathrm{T}}_{1}, ..., z_{i}\bar{z}^{\mathrm{T}}_{n}] = [s_{i,1},...,s_{i,n}],  
\label{eq:ps_sim}  
\end{aligned}
\end{equation}
where $s_{i}\in\mathbb{R}^{n}$ is a PS vector containing the PS between an image $x_{i}$ and all other images, and $s_{i,j}$ denotes the PS between images $x_{i}$ and $x_{j}$. $\mathrm{T}$ indicates that a matrix or a vector is transposed. Since all features are normalised as a unit vector, the computing PS is equivalent to the computing cosine angular similarity between two features:
\begin{equation}
s_{i,j}=z_{i}\cdot{}\bar{z}_{j} \equiv \lVert{}z_{i}\rVert{} \lVert{}\bar{z}_{j}\rVert{}cos\theta_{i,j} \equiv cos\theta_{i,j}.
\label{eq:sim_cos}
\end{equation}

Accordingly, all similarities are mapped into $[-1,1]$. When $s_{i,j}$ is closed to 1, then it is highly likely that two images (i.e., $x_{i}$ and $x_{j}$) show the same vehicle. On the other hand, if $s_{i,j}$ is closed to -1, then the two images probably represent different vehicles.

However, finding positive labels using only PS can generate plenty of false-positive results because there are a great number of image variations generated by illumination conditions of day and night, weather and climate changes, complex backgrounds. Hence, we propose a Dictionary-based Positive Label Mining (DPLM) for finding more reliable positive and negative label sets represented by
\begin{equation}
\text{P}^{+}_{i} = \text{DPLM}(y_{i},\mathcal{D}),
\label{eq:dfm}
\end{equation}
where $\text{DPLM}(\cdot)$ denotes the proposed DPLM module, and $\text{P}^{+}_{i}$ indicates a positive label set corresponding to an image $x_{i}$, respectively. The DPLM module is non-parametric; therefore, it is not affected by the model complexity.

The selected positive and negative labels are applied to the proposed Dictionary-based Triplet Loss (DTL) to optimise the embedding function $\phi$. Computing DTL is represented by  
\begin{equation}
\mathcal{L}_{\text{DTL}} = \sum_{m=1}^{B}\text{SSML}(\phi{}(x_{m}),\text{P}^{+}_{m},\text{N}^{-}_{m}),
\label{eq:dtl}
\end{equation}
where $\text{SSML}(\cdot)$ denotes the proposed SSML module, and $B$ indicates the training batch size. $\text{N}^{-}_{m}$ means a negative label set corresponding to an image $x_{m}$.

In the test step, with a given query image $q$ and a gallery $\mathcal{G}=\{g_{i}\}^{i=1:m}$, where $m$ is the number of images on $\mathcal{G}$, vehicle Re-ID process is conducted as follows:
\begin{equation}
\hat{g} = \arg\min_{g_{i} \in \mathcal{G}} dist(\phi(q),\phi(g_{i})), 
\label{eq:Re-ID}
\end{equation}
where $\hat{g}$ is a retrieved image possibly containing the same vehicle as $q$ among images in the gallery. $dist(\cdot)$ indicates a distance metric. 

Notably, in the training step, the elements in $\mathcal{D}$ have to be renewed to reflect the characteristic transformation of extracted features alongside parameter optimisation. Each element of $\mathcal{D}$ at $t$-th training step is updated by the average of the feature and the corresponding element on $\mathcal{D}$ as follows:
\begin{equation}
\bar{z}^{t}_{i} \xleftarrow{} \frac{\bar{z}^{t-1}_{i}+z^{t}_{i}}{2},  \quad \text{s.t. } \bar{z}^{0}_{i} = {z}^{0}_{i},
\label{eq:dic_update}
\end{equation}
where $z^{t}_{i}$ denotes the features extracted from image $x_{i}$ at the $t$-th training step, and $\bar{z}^{t-1}_{i}$ indicates the corresponding elements of $z^{t}_{i}$ stored in $\mathcal{D}$, respectively. To improve the robustness of $\mathcal{D}$, each element is regularised using $l2$-normalisation. However, the above scheme only can update a few elements of $\mathcal{D}$, and this partial update can be a cause of covariate shift problem \cite{he2016deep,schneider2020improving}. The covariate shift can negatively affect the model optimisation during model training. Therefore, to reduce the risk of the covariate shift, $\mathcal{D}$ is fully re-initialised at a specific point in the training step.

The proposed SSML based on DPLM and DTL improves the performance of unsupervised vehicle Re-ID by gradually improving the performance of label mining of DPLM and the discriminative power of learnt features iteratively. The details about the proposed DPLM and DTL are described in the following sections.

\subsection{Dictionary-based Positive Label Mining}
DPLM aims to find a high-quality positive label set that points out images showing the vehicle the same as the image $x_{i}$ among the thousands of unlabelled vehicle images. As aforementioned, the PS $s_{i}$ (see Eq. \eqref{eq:ps_sim}) may be insufficient as a criterion to distinguish highly reliable positive and negative label sets. 

To improve these label mining accuracy, we are filtering entire labels based on ${s}_{i}$  using a threshold $\tau$ to improve the mining performance as follows:
\begin{equation}
\text{P}^{\text{PS}}_{i} = \left\{ j ~|~ {s}_{i,j}\geq{}\tau, 1 \leq j \leq n \right\},
\label{eq:first} 
\end{equation}
where $\text{P}^{\text{PS}}_{i}$ denotes the label set containing features of positive label candidates for the image $x_{i}$. This process can improve the quality of label mining and reduce the computational cost by excluding the unnecessary labels for comparison. 

To improve the mining performance, we present a double-check scheme utilising \textit{Relative-Rank Consistency} and \textit{Adjacent Feature Distribution Similarity}. 
We assume that if two images belong to the same class, then the adjacent feature distributions of them should also be similar in the latent feature space.
In other words, two images should be a mutual neighbour for each other if they can be assigned as similar labels. This assumption is inspired by k-reciprocal nearest neighbour \cite{zhong2017re}. 

\textbf{Relative-rank consistency:} For a each index $j\in{}\text{P}^{\text{PS}}_{i}$, DPLM computes $s_{j}$ and finds out $\text{P}^{\text{PS}}_{j}$ containing the labels of top-$\text{K}$ nearest sample of $x_{j}$ with Eq.~\eqref{eq:first}. $\text{K}$ is defined by the cardinality of $\text{P}^{\text{PS}}_{i}$ as follows:
\begin{equation}
\text{K} \longleftarrow |\text{P}^{\text{PS}}_{i}|.
\label{eq:k} 
\end{equation}

If index $i$ is also one of the top-$\text{K}$ nearest labels of $j$, $j$ is considered as a positive label for $x_i$, decided by the rank-cyclic consistency check.  Above relative-rank consistency is inspired by Wang \etal \cite{wang2020unsupervised}. The set of the top-$\text{K}$ nearest feature labels of $j$-th element is defined by
\begin{equation}
\begin{aligned}
&\text{P}^{\text{PS}}_{j} = \mathop{\arg \operatorname {sort}}_t s_{j,t},\ \text{w.r.t.,}\ 1\leq{}t\leq{}n,\\
&\bar{\text{P}}^{\text{PS}}_{j} \longleftarrow{} \text{P}^{\text{PS}}_{j}[1:\text{K}],
\label{eq:s} 
\end{aligned}
\end{equation}
where $\bar{\text{P}}^{\text{PS}}_{j}$ is the filtered set as top-$K$ nearest labels of $\text{P}^{\text{PS}}_{j}$.
Note that the elements in $\text{P}^{\text{PS}}_{j}$ are sorted in descending order.

As a result, with a given image $x_{i}$, the positive label set, decided by the relative-rank consistency, is defined as follows:
\begin{equation}
\text{P}^{\text{Rank}}_{i} = \{\ j\ | \  \text{if} \ i\in\bar{\text{P}}^{\text{PS}}_{j} \ \text{w.r.t.,} \ j \in \text{P}^{\text{PS}}_{i}\},
\end{equation}
where $\text{P}^{\text{Rank}}_{i}$ defines the positive label set determined by the relative-rank consistency.

\textbf{Adjacent feature distribution similarity}: However, the relative-rank consistency only considers highly abstracted information (i.e., similarity ranking), 
and it can not comprehensively use other useful information such as the geometric distribution of neighbour features. Therefore, we present an adjacent feature distribution similarity to improve the quality of the mining results. 

To measure the adjacent feature distribution similarity, we first compute feature similarity distribution matrix $\hat{\mathcal{S}}\in\mathbb{R}^{n\times{}n}$ using the dictionary $\mathcal{D}$ as follows:
\begin{equation}
\hat{\mathcal{S}}=\mathcal{D} \mathcal{D}^{\mathrm{T}}=
\begin{bmatrix}
\bar{z}_{1}\cdot{}\bar{z}_{1}, \cdots, \bar{z}_{1}\cdot{}\bar{z}_{n}\\
\vdots \quad \quad \quad \ddots \quad \quad \quad \vdots \\
\bar{z}_{n}\cdot{}\bar{z}_{1}, \cdots, \bar{z}_{n}\cdot{}\bar{z}_{n}
\end{bmatrix}\equiv
\begin{bmatrix}
s_{1}\\
\vdots \\
s_{n}
\end{bmatrix}.
\label{eq:estimation}
\end{equation}
The $i$-th row vector of $\hat{\mathcal{S}}$ is equivalent to the PS vector of $x_{i}$. To remove the features that are positioned far from the target feature, we assign 0 to the elements that are lower than $\tau$ (i.e., $\forall s_i < \tau$) as same as Eq. \eqref{eq:first}.

Intuitively, $s_{i,j}$ indicates 
the similarity between the images $x_{i}$ and $x_{j}$ in the latent feature space;
therefore, $s_{i}$ can contain information about all neighbour features centred on $x_{i}$. We compute the adjacent feature distribution similarity $a_{i,j}$ between the two features $z_{i}$ and $z_{j}$ as follows:
\begin{equation}
\begin{aligned}
a_{i,j} = \lVert{}s_{i}-s_{j}\rVert{}, 
\label{eq:ads} 
\end{aligned}
\end{equation}
where 
$\lVert{}\cdot{}\rVert{}$ indicates Euclidean distance. 

The positive label sets based on the adjacent feature distribution similarity can be found by sorting the elements on $a_{i}$ in descending order and selects top-$\text{K}$ features as follows:
\begin{equation}
\begin{aligned}
&\text{Q}_{i} = \mathop{\arg \operatorname {sort}}_j a_{i,j}, \quad \text{w.r.t.,} 1 \leq j \leq n\\
&\text{P}^{\text{Adj}}_{i} \leftarrow{} \text{Q}[1:\text{K}],
\label{eq:second} 
\end{aligned}
\end{equation}
where $\text{Q}_{i}$ is the sorted labels based on $a_{i}$, and $\text{P}^{\text{Adj}}_{i}$ represents a positive feature set as the top-$\text{K}$ elements in $\text{Q}_{i}$. 

By considering $\text{P}^{\text{Rank}}_{i}$ and $\text{P}^{\text{Adj}}_{i}$ simultaneously, DPLM determines the true positive label set finally. A label is categorised as the true positive label set if it appears at both sets $\text{P}^{\text{Rank}}_{i}$ and $\text{P}^{\text{Adj}}_{i}$ at the same time. 
Otherwise, it is categorised as the negative label set. Intuitively, the duplicate elements mean that they are positioned at a near distance and have more mutual neighbours. The true positive label set $\text{P}^{+}_{i}$ is represented as
\begin{equation}
\text{P}^{+}_{i}= \{\ j\ |\ j\in\text{P}^{\text{Rank}}_{i} \wedge \text{P}^{\text{Adj}}_{i}\  \}
\quad \text{w.r.t.}~ 1 \leq j \leq n.
\label{eq:mlp} 
\end{equation}

\begin{figure}[t]
	\centering
	\includegraphics[width=\columnwidth]{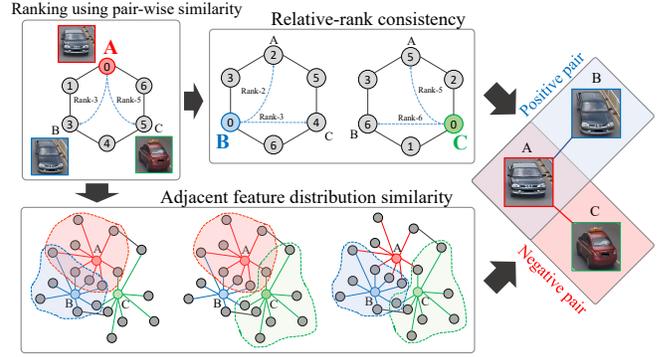}
	\caption{An illustration of the positive label mining using the relative-rank consistency and the adjacent feature distribution similarity on DPLM}
	\label{fig:2}
	\vspace{-2ex}
\end{figure}

Fig. \ref{fig:2} shows a conceptual illustration of the positive feature mining on DPLM. We demonstrate the effectiveness of the DPLM for unsupervised vehicle Re-ID on the ablation study in Section \ref{sec:exp:3}. 

\subsection{Dictionary-based Triplet Loss} 
For vehicle Re-ID, metric learning approaches have been applied based on a supervised way \cite{antonio2018unsupervised,chu2019vehicle}. 
Those methods were trained for minimising the distance between a given image and positive images and maximising the distance between the given image and negative images at the same time.
A traditional approach for solving the distance min-max problem is triplet loss formulated as
\begin{equation}
\mathcal{L}_{\text{Triplet}} =\lVert{}\phi(x^{a})-\phi(x^{p})\rVert{}-\lVert{}\phi(x^{a})-\phi(x^{p})\rVert{}+m,
\label{general_triplet} 
\end{equation}
where $x^{a}$, $x^{p}$, and $x^{n}$ indicate an anchor image (i.e., a given image), a positive image, and a negative image, respectively. $m$ denotes a pre-defined margin for the triplet loss. 

DPLM grants a capability to distinguish positive and negative samples for a given image and makes it possible to apply the triplet loss in an unsupervised way. However, there are two challenging issues in applying the triplet loss. First, for an input sample, commonly used metric learning takes one triplet inputs composed of three file paths to load image files for anchor $x^{a}$, positive sample $x^{p}$, and negative sample $x^{n}$. It needs additional processes to package the positive and negative samples and re-extract features from them. Second, the mining result is an imbalance composed of many negative samples and a few positive samples. Therefore, a trained model may suffer from the model collapse problem that a model is trained in the wrong direction (\eg biased into negative labels \cite{wang2020unsupervised}). We hence propose DTL to address those issues.

The DTL consists of two terms, $\mathcal{L}^{+}$ and $\mathcal{L}^{-}$, to minimise distances between a given image and positive labelled images and to maximise distances between the given image and the negative labelled images, respectively. With a given image $x_{i}$ and the positive label set $\text{P}^{+}_{i}$, the loss term for the positive features are defined as follows:
\begin{equation}
\begin{aligned}
\mathcal{L}^{+}(x_{i},\text{P}^{+}_{i}) &= \sum_{j\in{}\text{P}^{+}_{i}}\lVert{}s_{i,j}-1\lVert{}^{2},
\label{eq:positive} 
\end{aligned}
\end{equation}
where $\mathcal{L}^{+}$ denotes the loss term for $\text{P}^{+}_{i}$. Minimising $\mathcal{L}^{+}$ converged the distance between the latent feature $z_{i}$ and the stored features on the dictionary $\mathcal{D}$ selected by $\text{P}^{+}_{i}$ to 1, which can be interpreted as the highest cosine similarity value. 

On the other hand, the loss term for the negative features is defined as follows:
\begin{equation}
\begin{aligned}
\mathcal{L}^{-}(x_{i},\text{N}^{-}_{i}) &= \sum_{j\in{}\text{N}^{-}_{i}}\lVert{}\phi(x_{i})\cdot{}\bar{z}_{j}+1\lVert{}^{2},
\label{eq:negative} 
\end{aligned}
\end{equation}
where $\mathcal{L}^{-}$ denotes the loss term for $\text{N}^{-}_{i}$. Minimising $\mathcal{L}^{-}$ lead to converged the distance between $x_{i}$ and the negative labelled images into -1, which can be interpreted as the lowest cosine similarity value. 

Consequently, The DTL is defined by the summation of Eq. \eqref{eq:positive} and Eq. \eqref{eq:negative} as follows:
\begin{equation}
\begin{aligned}
\mathcal{L}_{\text{DTL}}(x_{i},\text{P}^{+}_{i},\text{N}^{-}_{i}) = \mathcal{L}^{+}(x_{i},\text{P}^{+}_{i}) + \sigma \mathcal{L}^{-}(x_{i},\text{N}^{-}_{i}),
\label{eq:loss_dtl} 
\end{aligned}
\end{equation}
where $\sigma$ denotes the balancing weight between the positive and negative loss terms. Compared with the general triplet loss (Eq. \eqref{general_triplet}), the proposed DTL is computationally less intensive because no extra process is needed to group the positive and negative samples and re-extract features for those samples.
 
\textbf{Hard-negative label mining}: Considering all negative labels is still problematic since there is a quantitative unbalance between positive and negative labels. To end this, DPLM includes a hard negative label mining scheme to select more informative negative labels for metric learning. As Eq. \eqref{eq:mlp}, the negative label set $\text{N}^{-}_{i}$ is defined by
\begin{equation}
\text{N}^{-}_{i} = \{j|j\notin \text{P}^{\text{Rank}}_{i} \wedge \text{P}^{\text{Ajd}}_{i} \quad \text{w.r.t.,} \ 1 \leq j \leq n\}. 
\label{eq:condition_for_neg} 
\end{equation}

The hard negative labels can be interpreted as vehicle images that look similar to a query image, but they actually represent different vehicles. Hence, we sort the negative label set $\text{N}^{-}_{i}$ in descending order by the PS $s$ and select top-$\gamma$\% of negative labels as the hard negative labels as follows: 
\begin{equation}
\begin{aligned}
&\text{N}^{-}_{i} = \mathop{\arg \operatorname {sort}}_{j\in\text{N}^{-}_{i}} s_{i,j},\\
&\bar{\text{N}}^{-}_{i} \leftarrow{} \text{N}^{-}_{i} [1:\lceil\gamma|\text{N}^{-}_{i}|\rceil],
\label{eq:mining_hard_label} 
\end{aligned}
\end{equation}
where $\bar{\text{N}}^{-}_{i}$ are the set for the hard negative labels. $|\text{N}^{-}_{i}|$ is the cardinality of $\text{N}^{-}_{i}$, and $\lceil\cdot\rceil$ denotes ceiling function. 

Consequently, with the given positive labels $\text{P}^{+}_{i}$ and the selected hard negative labels $\bar{\text{N}}^{-}_{i}$, the proposed method using DTL optimises the follow:
\begin{equation}
\begin{aligned}
\mathcal{L}_{\text{DTL}}(x_{i},\text{P}^{+}_{i},\bar{\text{N}}^{-}_{i}) = \mathcal{L}^{-}(x_{i},\text{P}^{+}_{i}) + \sigma \mathcal{L}^{-}(x_{i},\bar{\text{N}}^{-}_{i}).
\label{eq:min_max_representation} 
\end{aligned}
\end{equation}

In the proposed DTL, $\sigma$ and $\gamma$ can affect the Re-ID performance. $\sigma$ decides the weight about the loss term for the negative labels, and $\gamma$ decides the number of hard negative labels. A comprehensive ablation study about the proposed DTL and the effectiveness of $\sigma$ and $\gamma$ is performed in Section \ref{sec:exp:3}.

\begin{table}
\resizebox{\columnwidth}{!}{%
\begin{tabular}{l|c|c|c|c|c}
\hline
Dataset & Imgs &  Ids & Imgs$/$Id &  Cams & Time (H)\\
\hline\hline
VeRi-776 \cite{liu2016deep} & 49,360 & 776 & 63.60  & 18 & 18\\
\hline
VeRi-Wild \cite{wei2018person}&  416,314  &  40,671  &10.23 & 174 & 125,280\\
\hline
\end{tabular}
}
\caption{Key properties of the vehicle Re-ID datasets. `Imgs', `Ids', and `Cams' denote the number of images, identities, and cameras of each dataset, respectively.}
\vspace{-3ex}
\label{tbl:dataset_explain}
\end{table}

\section{Experiments}
\label{sec:exp}
\subsection{Dataset and Evaluation metrics}
\label{sec:exp:1}
Two publicly available datasets, VeRi-776 \cite{liu2016deep} and VeRi-Wild \cite{wei2018person}, are leveraged for ablation study and comparison with other methods. Key properties of the two datasets are represented in Table \ref{tbl:dataset_explain}. In particular, the VeRi-Wild dataset can be regarded as a more challenging dataset than the other dataset because the dataset is most recently proposed and suffers from substantial variations of scene conditions by illumination conditions of day and night and weather changes for a long time (30 days). Our experiments for unsupervised vehicle Re-ID have conducted with the standard protocols \cite{liu2016deep,liu2016deep_db,wei2018person}. Cumulative Matching Characteristics (CMC) and Mean Average Precision (mAP) are leveraged to evaluate the performances of unsupervised vehicle Re-ID methods.

\subsection{Implementation}
\label{sec:exp:2}
All images are resized to 256$\times$128, and all models are optimised using stochastic gradient descent (SGD) with a momentum of 0.9 for 60 epochs. The initial learning rates of the embedding function $\phi$ are 0.01. The learning rates are decayed by multiplying 0.1 for every 10 epoch, and the size of the batch is 256. ResNet-50 \cite{he2016deep} pre-trained by ImageNet \cite{krizhevsky2012imagenet} is adopted as the backbone of the embedding function $\phi$, and the dimensionality of the output for $\phi$ is 2,048. Simple data augmentations (such as random crop, rotation, and colour jitters) are used to improve the generalisation performance of learnt features. DPLM is conducted after 5 epochs to ensure the minimum label mining qualities. Before using the mining results, each feature is used as a positive label of itself. $\mathcal{D}$ is completely re-initialised every 5-epoch in the training step. The value of $\tau$ is set as 0.6. The balancing weight $\sigma$ and the hard negative mining rate $\gamma$ are fixed to 0.2 and 0.01, respectively, for the best performance (based on the ablation study). We implement our method using Pytorch, and all experiments are carried out with GTX TITAN RTX.

\begin{figure}[t]
    \centering
    \begin{subfigure}{0.49\linewidth}
        \includegraphics[width=\linewidth]{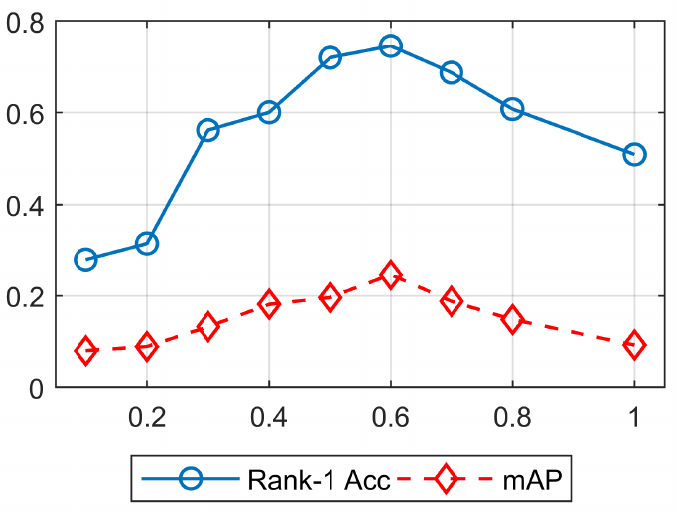}
    \caption{Experimental results on $\tau$}
    \end{subfigure}
    \begin{subfigure}{0.49\linewidth}
        \includegraphics[width=\linewidth]{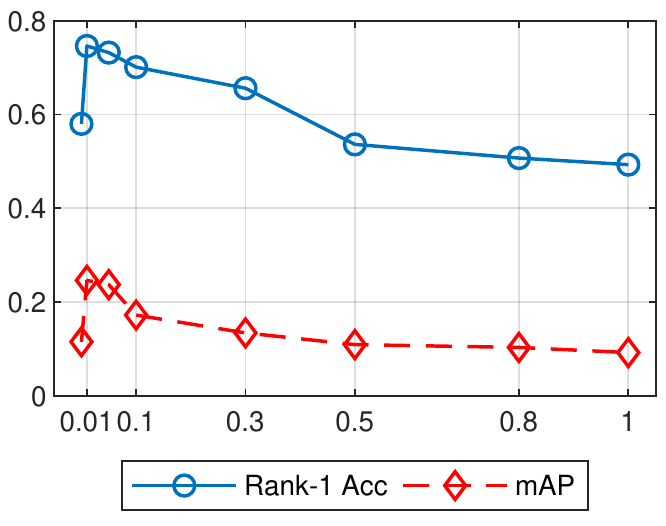}
    \caption{Experimental results on $\gamma$}
    \end{subfigure}
    \caption{Ablation studies for the $\tau$ in DPLM and the $\gamma$ and $\sigma$ in DTL. (a) and (b) represents the trends of rank-1 accuracies and mAPs with respect to the value of $\tau$ and $\gamma$ on VeRi-776 datasets.}
\label{fig:as_tau_gamma}   
\end{figure}

\begin{table}
\resizebox{\columnwidth}{!}{%
\begin{tabular}{r||c|c|c|c|c|c}
\hline
$\sigma$ setting & 0.01 & 0.1 & 0.2 & 0.4 & 0.8 & 1.0  \\
\hline
Rank-1 Acc. & 59.3 & 69.2 & 72.7 & \textbf{74.1} & 67.3 & 55.1 \\
\hline
mAP & 15.2  & 19.3 & 22.1 & \textbf{26.7}  & 18.6 & 14.2  \\
\hline
\end{tabular}
}
\caption{Performance analysis depending on the setting of $\sigma$. The \textbf{bolded} figures denote the best performance.} 
\vspace{-2ex}
\label{tbl:comparison-sigma}
\end{table}

\subsection{Ablation Study}
\label{sec:exp:3}
We evaluate the proposed unsupervised vehicle Re-ID performance depending on the setting of $\tau$, $\gamma$, and $\sigma$. Also, we demonstrate the effectiveness of DPLM and DTL. All experiments are conducted with unsupervised vehicle Re-ID settings on the VeRi-776 dataset. Parameters that are not subject to monitoring are fixed during the experiments.

\textbf{Parameter analysis on $\tau$}: The $\tau$ affects the quality of the positive label mining on DPLM. As shown in Fig. \ref{fig:as_tau_gamma}(a), the curves of the rank-1 accuracy and mAP are rapidly grown-up until $\tau$ of 0.5 and start to diminish after 0.7 of $\tau$ clearly. Those trends can be interpreted as follows. When $\tau$ is too low, many false-positive labels are predicted. On the other hand, when $\tau$ is too high, the amount of predicted positive labels is not enough to cover various representations of the vehicle. The best performance is achieved by $\tau$ of 0.6, and this value would be fixed in further experiments.

\textbf{Parameter analysis on $\gamma$}: Fig. \ref{fig:as_tau_gamma}(b) shows that too large or too small values of $\gamma$ negatively affect the vehicle Re-ID performance. Too large $\gamma$ is harming the model performance. A low $\gamma$ will degrade the model performance since it may insufficient to make a margin between positive features and negative features on complex feature distribution. The best performance is achieved by $\gamma$ of 0.01, and this value would be fixed in further experiments.

\textbf{Parameter analysis on $\sigma$}: Table \ref{tbl:comparison-sigma} reports the analysis of the balancing weight $\sigma$ of DTL. $\sigma=1.0$ means that there is no bias between the positive term (Eq. \eqref{eq:positive}) and the negative term (Eq. \eqref{eq:negative}) in the model training. The experimental results in Table \ref{tbl:comparison-sigma} show that when $\sigma=1.0$, our model can not produce the gradients to pull positive labels together. When $\sigma=1.0$, the rank-1 accuracy and mAP are 55.1 and 14.2, respectively, and these figures are the lowest figures on our experiments. When $\sigma$ becomes smaller, the performance is being improved; however, too small $\sigma$ makes performance degradation. We set $\tau$ as 0.6 in further experiments. 

\begin{figure}[t]
	\centering
	\includegraphics[width=\columnwidth]{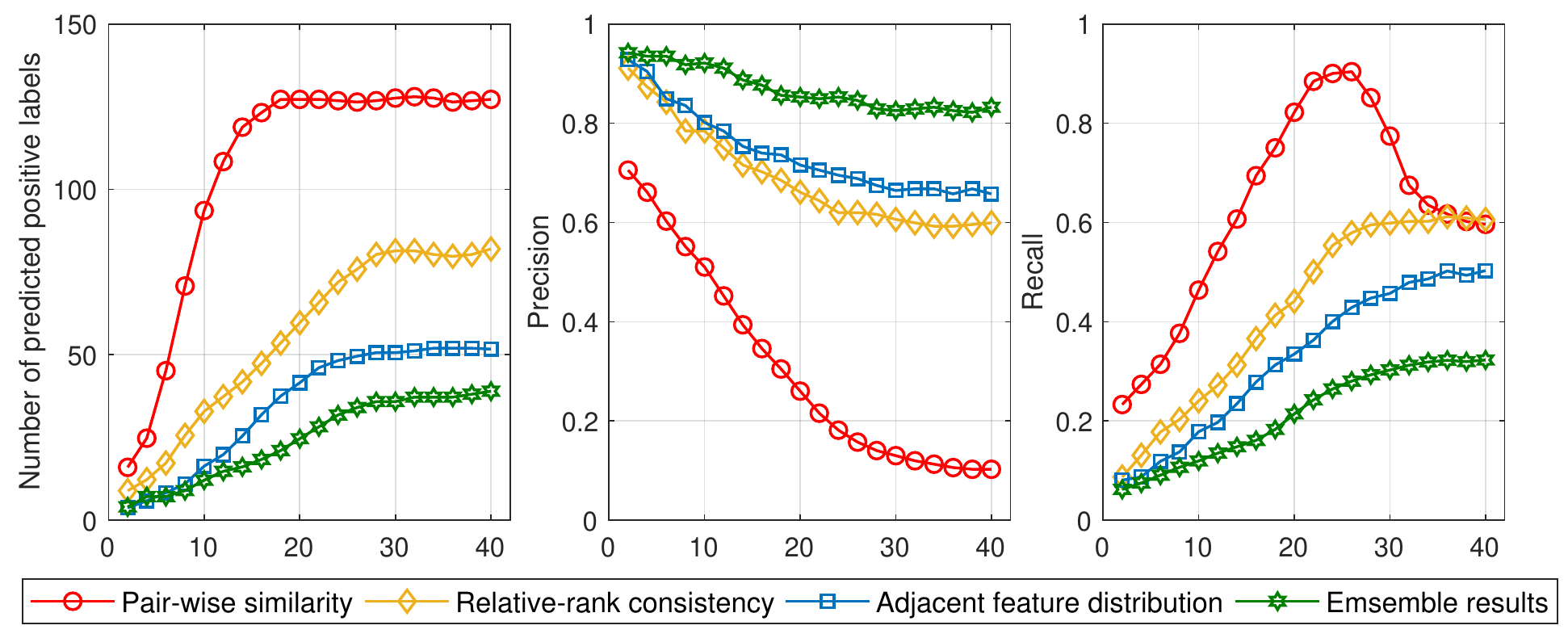}
	\caption{Analysis of the positive label mining performances based on the pair-wise similarity $\text{P}^{\text{PS}}$, the relative-rank consistency $\text{P}^{\text{Rank}}$, the adjacent feature distribution similarity $\text{P}^{\text{Adj}}$, and the ensemble of these metrics $\text{P}^{+}$. The amount of predicted positive labels, precision, and recall are shown with respect to the training epoch.}
	\label{fig:mul_performance}
\end{figure}

\textbf{Effectiveness of DPLM}: DPLM predicts positive labels with four metrics: 1) pair-wise similarity $\text{P}^{\text{PS}}$, 2) relative-rank consistency $\text{P}^{\text{Rank}}$, 3) adjacent feature distribution similarity $\text{P}^{\text{Adj}}$, and 4) ensemble of those metrics $\text{P}^{\text{+}}$. We compare the positive label mining performance of those metrics. As shown in Fig. \ref{fig:mul_performance}, $\text{P}^{\text{PS}}$ achieves the highest recall curve, but its precision is extremely lower than that of others because a number of positive labels with false-positives are too many. 
The precision curves of $\text{P}^{\text{Rank}}$ and $\text{P}^{\text{Adj}}$ are lower than the ensemble results $\text{P}^{\text{+}}$ of those two metrics. However, the recall of $\text{P}^{\text{+}}$  is the lowest among the four metrics since the number of predicted labels is the smallest. The above results can be interpreted that the positive label prediction accuracy is more important than the amount of predicted positive labels in improving unsupervised vehicle Re-ID performance.

\begin{table}
\resizebox{\columnwidth}{!}{%
\begin{tabular}{c||c||c|c|c}
\hline
\multirow{2}{*}{Loss function} & \multirow{2}{*}{Positive Labels} & \multicolumn{3}{c}{VeRi-776}  \\
\cline{3-5} 
& & Rank-1  & Rank-5  & mAP     \\
\hline
\multirow{4}{*}{Triplet Loss ($\mathcal{L_{\text{Triplet}}}$)} & $\text{P}^{\text{PS}}$ & 32.8 & 53.3 & 6.2 \\
& $\text{P}^{\text{Rank}}$  & 65.7 & 73.6 & 16.31  \\
& $\text{P}^{\text{Adj}}$  & 63.2 & 75.1 & 15.62  \\
\cline{2-5}
& $\text{P}^{+}$ & 70.9 & 78.9 & 20.6 \\
\hline\hline
\multirow{4}{*}{DTL ($\mathcal{L}_{\text{DTL}}$)} & $\text{P}^{\text{PS}}$ & 41.6 & 61.7 & 9.2 \\
&$\text{P}^{\text{Rank}}$  & 71.9 & 78.6 & 22.9 \\
&$\text{P}^{\text{Adj}}$  & 72.3 & 78.3 & 23.5 \\
\cline{2-5}
&$\text{P}^{+}$ &  \textbf{74.5} & \textbf{80.3} & \textbf{26.7} \\
\hline
\end{tabular}
}
\caption{Quantitative performance comparison of loss functions and positive label mining approaches} 
\vspace{-2ex}
\label{tbl:comparison-loss&mlp}
\end{table}

\begin{table*}[t]

\begin{center}
\resizebox{\textwidth}{!}{  
\setlength{\tabcolsep}{3pt}
\begin{tabular}{l|c| c|c| c c c |c| c c c| c  c c|c c c}
\hline
\multirow{2}{*}{Methods} & \multirow{2}{*}{Year} & \multirow{2}{*}{Settings} & \multirow{2}{*}{Source}  & \multicolumn{3}{c|}{VeRi-776} &  \multirow{2}{*}{Source}  &  \multicolumn{3}{c|}{VeRi-Wild (Small)} & \multicolumn{3}{c|}{VeRi-Wild (Medium)} & \multicolumn{3}{c}{VeRi-Wild (Large)}\\
 \cline{5-7} \cline{9-17}
& & & & Rank-1  & Rank-5  & mAP & & Rank-1& Rank-5 & mAP  & Rank-1 & Rank-5 & mAP & Rank-1  & Rank-5  & mAP \\
\hline\hline
SPGAN \cite{deng2018image} & 2018  & DA & VehicleID  & 57.4 & 70.0 & 16.4 & VehicleID  &     59.1     &     76.2     &     24.1     &     55.0     &   74.5       &    21.6      &     47.4     &     66.1     &    17.5  \\
VR-PROUD \cite{bashir2019vr} & 2019 & DA & VehicleID &  55.7 & 70.0 & 22.7 & - & - & - & - & - & - & - & - & - & - \\
ECN \cite{zhong2019invariance} & 2019 & DA &  VehicleID & 60.8 & 70.9 & 27.7 & VehicleID  &   73.4       &     88.8     &     34.7     &   68.6       &     84.6     &    30.6      &   61.0       &     78.2     &    24.7 \\
PAL \cite{he2019part}& 2020 & DA & VehicleID &  68.2 &  79.9 & \underline{42.0} & - & - & - & - & - & - & - & - & - & - \\
UDAP \cite{song2020unsupervised}& 2020 & DA & VehicleID & 76.9 & \underline{85.8} & 35.8 & VehicleID &     68.4    &     85.3     &    30.0      &    62.5      &     81.8     &    26.2      &     53.7     &       73.9 &     20.8  \\
VACP-DA \cite{zheng2020aware}& 2020 & DA & VehicleID &   \underline{77.4} & 84.6 & 40.3 & VehicleID & \underline{75.3} & \underline{89.0} & \underline{39.7} & \underline{69.0} & \underline{85.5} & \underline{34.5} & \underline{61.0} & \underline{79.7} & \underline{27.4} \\
AE \cite{journals/tomccap/DingFXY20} & 2020  & DA & VehicleID & 73.4 & 82.5 & 26.2 & VehicleID  &     68.5     &     87.0     &  29.9        &     61.8     &     81.5     &   26.2       &     53.1     &    73.7      &    20.9  \\
\hline\hline
LOMO$^{\ddagger}$ \cite{liao2015person} & 2015  & Uns &  - &  42.1 & 62.2 & 12.2 & -  & 25.7  & 44.7  & 8.9 & 23.6 & 40.6  &   8.1 &  18.8 & 34.4& 5.9  \\ 
BOW$^{\ddagger}$ \cite{zheng2015scalable} & 2015  & Uns & -     & 44.7    & 66.4  &  14.5 & -   &    28.5     &   43.6       &     9.4     &     25.4     &     40.7     &    8.6      &     18.3    &    38.6      &   6.6 \\
BUC$^{\ddagger}$ \cite{lin2019bottom}& 2019  & Uns & -     & 54.7    & 70.4  &  21.2 & -   &     37.5     &   53.0       &     15.2     &     33.8     &     51.1     &    14.8      &     25.2     &    41.6      &   9.2 \\
SSL$^{\ddagger}$ \cite{lin2020unsupervised} & 2020  & Uns & -     & 69.3    & 72.1  &  23.8 & -   &     38.5     &   58.1       &     16.1     &     36.4     &     56.0     &    17.9      &     32.7     &    48.2      &   13.6 \\
MMLP$^{\ddagger}$ \cite{wang2020unsupervised} & 2020  & Uns & -     & 71.8    & 75.9  &  24.2 & -   &     40.1     &   63.5       &     15.9     &     39.1     &     60.4     &    19.2      &     33.1     &    50.4      &   14.1 \\
\hline \hline
Ours & 2021 & Uns &  - &  \textbf{74.5} & \textbf{80.3} & \textbf{26.7} & - & \textbf{49.6} & \textbf{71.0} & \textbf{23.7} & \textbf{43.9}  & \textbf{64.9} & \textbf{20.4} & \textbf{34.7} & \textbf{55.4} & \textbf{15.8}\\
\hline
\end{tabular}
}
\end{center}
\caption{Performance comparison on unsupervised vehicle Re-ID with state-of-the-art methods, including different unsupervised Re-ID domains such as person Re-ID, in terms of rank-1 and rank-5 accuracies and mAP on the VeRi-776 dataset~\cite{liu2016deep} and the VeRi-Wild dataset~\cite{wei2018person}. `-' denotes that the results are not provided, and `$\ddagger$' denotes that the results are obtained by our experiments using publicly available source code. The underlined results indicate the best performance among the DA-based methods. The \textbf{bolded} results indicate the best performance on unsupervised methods.}
\label{tbl:comparison}
\end{table*}

\textbf{Effectiveness of DTL}: DTL is proposed to provide a stable metric learning process using the outcome of DPLM. We train our method using the general triplet loss (Eq. \eqref{general_triplet}) and DTL with the four types of multi-labels $\text{P}^{\text{PS}}$, $\text{P}^{\text{Rank}}$, $\text{P}^{\text{Adj}}$, and $\text{P}^{+}$. Table \ref{tbl:comparison-loss&mlp} contains the rank-1 and rank-5 accuracies and the mAP of the two loss metrics. The experimental results show that DTL can provide better performance than the general triplet loss. Entire experimental results on Table \ref{tbl:comparison-loss&mlp} justify the effectiveness of DTL for unsupervised vehicle Re-ID.

\subsection{Comparison with state-of-the-art methods}
\label{sec:exp:4}
Our method is compared with various recent state-of-the-art unsupervised Re-ID methods. Unfortunately, on unsupervised vehicle Re-ID, a limited number of works reported performance for those two datasets, and even those methods are based on domain-adaptation \cite{he2019part,song2020unsupervised,zheng2020aware}. As a result, we additionally compare our method with the state-of-the-art methods for unsupervised person Re-ID \ie LOMO  \cite{liao2015person}, BOW \cite{zheng2015scalable}, OIM \cite{xiao2017joint}, BUC \cite{lin2019bottom}, SSL \cite{lin2020unsupervised}, and MMLP \cite{wang2020unsupervised}, which do not need any types of labelled dataset. Those methods released source code on public repositories, so the evaluation of those methods is conducted based on their source code.  

Table \ref{tbl:comparison} shows quantitative comparison on VeRi-776 and VeRi-Wild datasets. On the comparison, the methods (LOMO \cite{liao2015person} and BOW \cite{zheng2015scalable}) based on hand-crafted features produce lower performance than others. Our method outperforms BUC \cite{lin2019bottom} with large margins. This performance gap can be interpreted as follows. First, their performance can be variant to the positions of initial clusters. This issue is one of the inevitable issues for all clustering-based methods. Second, those methods do not consider the number of data assigned to each cluster, which means unbalance between each positive label can be caused.

Our method also achieves better performance than SSL \cite{lin2020unsupervised} and MMCL \cite{wang2020unsupervised}. Both methods initially assign the file indices as single-class labels, so those methods are methodologically similar to our work. The difference in performance can be interpreted as follows. In the training step, SSL only considers a pair-wise similarity of features, so it may create lots of false-positive results. MMCL takes into account a cycle consistency as similar to our relative-rank consistency, but DPLM on our method can check the consistency and the similarities of each features' neighbour simultaneously. As shown in Table \ref{tbl:comparison-loss&mlp}, the ensemble of the two criteria can improve the accuracy of positive label searching.

In the comparison with the methods based on DA, our method produces comparable performances to the state-of-the-art methods. For example, the rank-1 accuracy of our method on the VeRi-776 dataset is 74.5. The rank-1 accuracy of our method is clearly higher than that of several approaches \cite{bashir2019vr,he2019part}. On the experiments using the VeRi-Wild dataset, the rank-1 accuracy of our method is 49.6, 43.9, and 34.7 for the small, medium, and large test sets, respectively. Although our method achieves the best performance among the unsupervised approaches, still there is a margin between our method and the DA-based methods  \cite{deng2018image,zhong2019invariance,zheng2020aware,journals/tomccap/DingFXY20}. 

However, these performance gaps have been obtained by using transferred information from a labelled source dataset. Additionally, several studies leveraged various complementary features such as annotated camera views \cite{he2019part,song2020unsupervised} and vehicle part-awareness \cite{zheng2020aware}. These features would be particularly helpful for the VeRi-Wild dataset that contains a great number of image variation, occurred by illumination condition of day and night and weather condition. Consequently, our method can be considered as a more flexible solution because our method does not need any pre-labelled dataset. 

\section{Conclusion}
\label{sec:con}
This paper has proposed self-supervised metric learning (SSML) using a feature dictionary. The proposed method extracts feature through an embedding function from all images and stores them in the dictionary. Based on the feature dictionary, when a vehicle image is given, the dictionary-based positive label mining (DPML) searches positive labels by computing the feature's pairwise similarity, relative-rank consistency, and adjacent feature distribution similarity in each training step. The excavated positive labels are applied to the dictionary-based triplet loss (DTL) to improve the discriminative power of the embedding function and the quality of positive labels predicted by DPML. The results of ablation studies have demonstrated the effectiveness of DPML and DTL for unsupervised vehicle Re-ID. Compared with existing various state-of-the-art methods on unsupervised vehicle Re-ID, the proposed SSML has outperformed other unsupervised methods and shown the competitive results compared with the domain-adaptation based methods that need a pre-labelled dataset to train their models.

\small
\bibliographystyle{IEEEtran}
\bibliography{iros_bib}

\end{document}